# Minutiae Based Thermal Face Recognition using Blood Perfusion Data

Ayan Seal
Department of Computer Science and Engineering dept.
Jadavpur University
Kolkata-700032, India
ayan.seal@gmail.com

Mita Nasipuri
Department of Computer Science and Engineering dept.
Jadavpur University
Kolkata-700032, India
mnasipuri@cse.jdvu.ac.in

Debotosh Bhattacharjee
Department of Computer Science and Engineering dept.
Jadavpur University
Kolkata-700032, India
debotosh@indiatimes.com

Dipak Kumar Basu
Department of Computer Science and Engineering dept.
Jadavpur University
Kolkata-700032, India
dipakkbasu@gmail.com

*Abstract*— **This paper describes an efficient approach for human face recognition based on blood perfusion data from infra-red face images. Blood perfusion data are characterized by the regional blood flow in human tissue and therefore do not depend entirely on surrounding temperature. These data bear a great potential for deriving discriminating facial thermogram for better classification and recognition of face images in comparison to optical image data. Blood perfusion data are related to distribution of blood vessels under the face skin. A distribution of blood vessels are unique for each person and as a set of extracted minutiae points from a blood perfusion data of a human face should be unique for that face. There may be several such minutiae point sets for a single face but all of these correspond to that particular face only. Entire face image is partitioned into equal blocks and the total number of minutiae points from each block is computed to construct final vector. Therefore, the size of the feature vectors is found to be same as total number of blocks considered. For classification, a five layer feed-forward backpropagation neural network has been used. A number of experiments were conducted to evaluate the performance of the proposed face recognition system with varying block sizes. Experiments have been performed on the database created at our own laboratory. The maximum success of 91.47% recognition has been achieved with block size 8×8.**

I.  INTRODUCTION

Biometric is an emerging technology which utilizes vein scan, facial thermogram, DNA matching, blood pulse, ear shape, gait recognition etc in security systems for establishing the identity of a person. Much work is still needed for design of convenient, secure and privacy-friendly systems. The word biometric is derived from the ancient Greek words "bios" meaning life and "metron" meaning measure [1],[2]. So, the meaning of biometric is life measurement. Biometrics uses physical characteristic or personal trait to match with the data in the database to determine the possible candidates. Physical feature is suitable for identity verification and generally obtained from living human body. Commonly used physical features are fingerprints, facial features, hand geometry, and eye features (iris and retina) etc. Personal trait is more appropriate in application but with less secure. The most commonly used personal traits are signature and voice. Much work is still needed to improve the biometric security systems. There is a number of reasons to choose face recognition for designing efficient biometric security systems. The most important one is that no physical interaction is needed, it is the only biometric those allow you to perform passive detection in different circumstances. Since last three decades there exist many commercially available systems of face recognition technology to identify human faces; however face recognition is still a challenging area due to various facial expressions, poses, non-uniform light illuminations and occlusions. Mainly there are two methods of capturing an image. One is visual imaging and the other is thermal imaging. Visual images captured by optical cameras are more common than thermal images captured by infra-red cameras. Face recognition techniques have biased towards visual images due to easy availability of low cost visible band optical cameras. It, however, requires an external source of illumination. Recently researchers have been using near-infrared imaging camera for face recognition with good results [3]. The tasks of face detection, location, and segmentation are relatively easier and more reliable for thermal face images then their visual counterpart [4]. Thermal imaging has better accuracy as it uses facial temperature variations caused by vein structure on facial surface as the distinguishing trait. As the heat pattern is emitted from the face surface itself without any source of external radiation these systems can capture images despite the any external lighting and even in the dark. Humans are homoeothermic and hence capable of maintaining constant temperature under different surrounding temperature. A facial thermal pattern is determined by the vascular structure of each face, [5] which are unique. An infrared camera with good sensitivity can indirectly capture images of superficial blood vessels on the human face [6]. However, it has been indicated by Guyton and Hall [7] that the average diameter of blood vessels is around 10-15 μm, which is too small to be detected by current IR cameras because of the limitation in spatial resolution. The skin just above a blood vessel is on an average 0.1 ◦C warmer than the adjacent skin, which is beyond the thermal accuracy of current IR cameras. The convective heat transfer effect from the flow of "hot" arterial blood in superficial vessels creates characteristic thermal imprints, which are at a gradient with the surrounding tissue. Face recognition method does not depend only on the topology of the facial vascular network but also on the fat depositions and skin complexion. The reason is that imagery is formed by the thermal imprints of the vessels and not the vessels directly. Even if the vessel topology was absolutely the same across individuals still, the thermal imprints would differ due to variable absorption from different fat padding (skinny faces versus puffy faces) [8]. It has been found that there exists an analogy between thermal imprints of human faces and fingerprints of human beings. The thermal imprints of the blood vessels may be treated as the ridges in the fingerprints and fingerprints recognition techniques may be applied on thermal imprints of the human faces for their recognition. The reliability of the fingerprint recognition system has been significantly enhanced through the technique of minutiae extraction from ridges. Most common type of minutiae is: when a ridge either comes to an end, which is called a ridge-termination or when it splits into two ridges, which is called a ridge-bifurcation. Fig. 1a) illustrates an example of a ridge bifurcation and 1b) depicts an example of a ridge termination point.

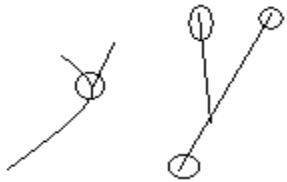

Fig. 1 a) Ridge bifurcation  b) Ridge termination.

So, the number and locations of the minutiae vary from face to face in any particular person. When a set of face images is obtained from an individual person, the number of minutiae is recorded for each face. The precise locations of the minutiae are also recorded, in the form of numerical coordinates, for each face and the resultant function thus obtained has been stored in a separate database. A computer can rapidly compare this function with that of anyone else in the world whose face image has been scanned. The paper is organized as follows: Section II presents about the physiological features of a thermal infra-red image. Section III describes the different steps of proposed approach. Section IV shows the experiment and results and finally, Section V concludes and remarks about some of the aspects analyzed in this paper.

## II. PHYSIOLOGICAL FEATURES OF A THERMAL FACE

A typical thermal infrared face image is shown in Fig. 3, which depicts interesting thermal information of a facial model. The temperature variation across the face can be easily visualized as the different colour regions in the thermogram. In Fig. 2 yellow colour represents average temperature of $97.3^0$F, bright green colour represents average temperature $95.1^0$F, brown colour represents average temperature of $93.7^0$F, blue colour represents average temperature of $89.6^0$F and pink colour represents average temperature of $91.4^0$F. Physiological features of a thermal face image have already been discussed in details in the introduction section. In this work, these characteristics have been used to build a unique thermal faceprint of a person.

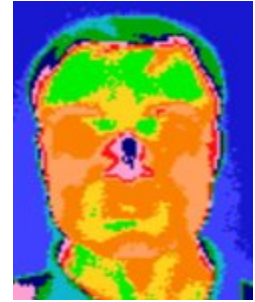

Fig. 2 A thermal infrared image.

## III. PROPOSED APPROACH

The first step in the proposed approach, as shown in Fig.3, is to extract the face skin region from grayscale thermal (infra-red) image. Initially, each of the captured 24-bits colour images have been converted into its 8-bit grayscale image counterpart. Then convert binary image from converted grayscale image. The resultant image replaces all pixels in the grayscale image with luminance greater than mean intensity with the value 1 (white) and replaces all other pixels with the value 0 (black). Different 8-connected objects present in 2-D binary image have been extracted [9] and the largest component among them has been identified as face skin region and other small components, which are other parts of the image, have been rejected. Fig. 4, depicted the largest component as a face skin region. There after crop the image is cropped and unwanted part of the image i.e. the background is eliminated. In binary image, black pixels mean background and white pixels mean the face region. Cropping process starts from top left corner and top right corner of the binary image along the lines and traverses parallel to vertical axis. This process stops when it encounters a white pixel first and then draw a vertical line from two points (one is left side of the face and another is right side of the face), eliminates the left part and right part (i.e. black pixel) of the lines respectively. In the same way it eliminates the upper and lower side of the face region. Fig. 5 shows the cropped image. Morphological erosion [9] is used to extract the thermal physiological face features and construct the region having constant or equal temperatures. Medial axis transform [9] is applied to extract lines similar to isothermal lines in weather maps linking all points of equal or constant temperature, in order to get blood perfusion data. Fig. 6 shows blood perfusion image of a thermal face image. The extracted blood perfusion image is nothing but border area of two different regions.

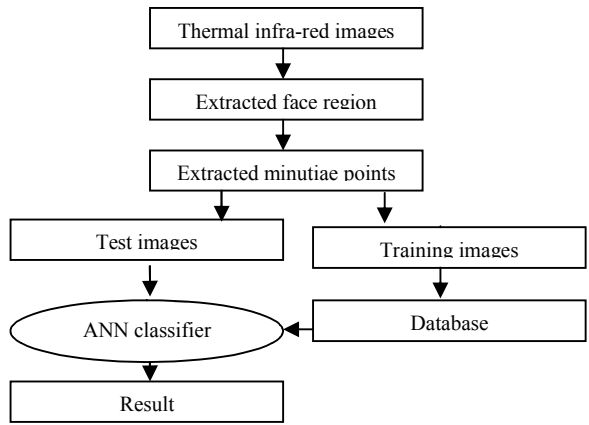

Fig. 3 Schematic block diagram of the proposed approach.

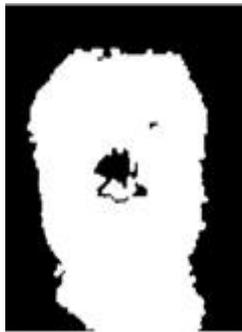

Fig. 4 The largest component as face skin region with black back ground.

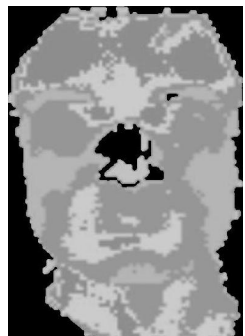

Fig. 5 Cropped face image.

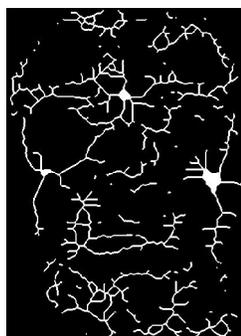

Fig. 6 Blood perfusion image

The concept of minutiae extraction from finger print recognition have been taken from [7],[10],[11] and applied to our work. Here fingerprint's ridges are like blood perfusion data of a face. The uniqueness of a face's blood perfusion data can be determined by the pattern of ridges as well as the minutiae points. Minutiae points are local ridge characteristics that occur either at a ridge bifurcation or at a ridge termination.

| 0 | 0 | 1 | | 1 | 1 | 0 | | 0 | 1 | 0 |
|---|---|---|---|---|---|---|---|---|---|---|
| 0 | 1 | 0 | | 1 | 1 | 0 | | 0 | 1 | 1 |
| 0 | 0 | 0 | | 0 | 0 | 0 | | 0 | 0 | 0 |
| a) | | | | b) | | | | c) | | |

Fig. 7 Binary number indicating the minutiae point.

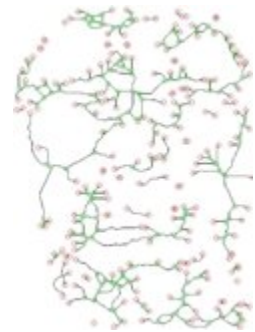

Fig. 8 Minutiae points.

The number of '1''s within each 3x3 window on the blood perfusion image where minutiae points are essentially the terminations and bifurcations of the ridge lines that constitutes a faceprint is computed. This is the vital part of the minutiae extraction of the faceprint image where the termination point and bifurcation point will be determined. If the central cell has a '1' and has another '1' as it's only neighbours, then it is a termination point like in Fig. 7a. If the central cell contains a '1' and has three '1''s as neighbours, then it is an bifurcation as shown in Fig. 7b and if the central cell is '1' and has two '1''s as neighbours, then it is a normal point like in Fig. 7c. Due to various noises in the face print image, the extraction algorithm produces a large number of spurious minutiae points. Therefore, differentiating spurious minutiae from real minutiae in the post-processing stage is crucial for accurate face recognition. The more spurious minutiae are eliminated; the better will be the classification performance. In addition, classification time will be significantly reduced because reduction of feature points. Thus the whole image is divided into number of fixed sized blocks. The size of the each block may be 8×8, 16×16, and 32×32. The number of minutiae points in each block have been counted and stored in a vector. Fig. 8 illustrates minutiae points, which are basically the bifurcation and termination points, using green and red colors respectively. For each face image, one corresponding vector has been found i.e. total number of such vectors is equal to total number of faces. Then divide these vectors into two sets one for training purpose and another for testing purpose. ANN classifier has been used to classify each of these vectors [12],[13]. Here a five layer feed-forward back propagation neural network has been used for this purpose. First hidden layer contain 100 neurons, second hidden layer contain 50 neurons, and third hidden layer holds 10 neurons and the last layer contain 6 neurons because 6 different persons are there in our experiment. Tan-sigmoid transfer functions is used to calculate a layer's output from its net the first input and the next three hidden layers and the outer most layer gradient descent with momentum training function is used to updates weight and bias values.

## IV. EXPERIMENT AND RESULTS

Experiments have been performed on our own thermal face images captured using a FLIR 7 camera. Some preprocessing has been done for the image database used in this paper. A typical thermal infrared face image is represented by different colour regions with different temperatures which, has been already discussed in details in the section II. All the training and testing images are grayscale images of size 416×544. Face images of 6 persons (each having 34 different images) were captured in normal temperature conditions till today. The obtained results are shown in Table 1 for different block sizes.

TABLE I
PERFORMANCE RATE FOR DIFFERENT BLOCK SIZE.

| No of Block | Performance Rate |
|---|---|
| 8×8 | 91.47% |
| 16×16 | 69.44% |
| 32×32 | 83.33% |

## V. CONCLUSION

A minutiae based thermal face recognition using blood perfusion data has been proposed here. Entire face image is divided into equal number of blocks and the total number of minutiae points from each block is considered as one feature. Features from all the blocks are combined to create the final feature vector. Classification of these feature vectors has been done using a multilayer perceptron. Final recognition rate has been enhanced by varying size of the blocks. One of the major advantages of this approach is the ease of implementation. Furthermore, no knowledge of geometry or specific feature of the face is required. However, this system is applicable to front views and constant background only. It may fail in unconstraint environments like natural scenes

ACKNOWLEDGMENT

Authors are thankful to a major project entitled "Design and Development of Facial Thermogram Technology for Biometric Security System," funded by University Grants Commission (UGC),India and "DST-PURSE Programme" at Department of Computer Science and Engineering, Jadavpur University, India for providing necessary infrastructure to conduct experiments relation to this work.